\documentclass{article}
\usepackage{spconf,amsmath,graphicx}
\usepackage{booktabs}
\usepackage{enumitem} 

\title{YODAS: YouTube-Oriented Dataset for Audio and Speech}
\name{
    \begin{tabular}[c]{@{}c@{}c@{}c@{}c@{}c@{}}
        Xinjian Li$^1$, Shinnosuke Takamichi$^2$,Takaaki Saeki$^1$, William Chen$^1$, Sayaka Shiota$^3$, Shinji Watanabe$^1$ \\
    \end{tabular}
}
\address{$^1$Carnegie Mellon University, USA\\
$^2$The University of Tokyo, Japan\\
$^3$Tokyo Metropolitan University, Japan
}

\begin{document}
%
\maketitle
\begin{abstract}

In this study, we introduce YODAS (YouTube-Oriented Dataset for Audio and Speech), a large-scale, multilingual dataset comprising currently over 500k hours of speech data in more than 100 languages, sourced from both labeled and unlabeled YouTube speech datasets. The labeled subsets, including manual or automatic subtitles, facilitate supervised model training. Conversely, the unlabeled subsets are apt for self-supervised learning applications. YODAS is distinctive as the first publicly available dataset of its scale, and it is distributed under a Creative Commons license\footnote{https://huggingface.co/datasets/espnet/yodas}
. We introduce the collection methodology utilized for YODAS, which contributes to the large-scale speech dataset construction. Subsequently, we provide a comprehensive analysis of speech, text contained within the dataset. Finally, we describe the speech recognition baselines over the top-15 languages.

\end{abstract}
\begin{keywords}
multilingual speech processing, speech recognition, large-scale speech dataset
\end{keywords}
\section{Introduction}
\label{sec:intro}

In recent years, significant advancements have been achieved in the field of speech recognition. With a sufficiently large speech dataset, it becomes feasible to train various end-to-end models using objectives such as CTC, ASG, seq2seq, RNN Transducer, and others~\cite{graves2006connectionist, collobert2016wav2letter, graves2013speech, sutskever2014sequence,prabhavalkar2023end}. We also observe the trend of using self-supervised learning models such as HuBERT and wav2vec2 to take advantage of unlabeled datasets~\cite{baevski2020wav2vec,hsu2021hubert,mohamed2022self}. Those improvements have been realized primarily through the utilization of large-scale multilingual speech datasets. For example, the BABEL project was a pioneering endeavor that scaled multilingual capabilities significantly~\cite{gales2014speech}. The Common Voice project, facilitates an online speech collection interface, offering speech datasets in over 100 languages and encompassing 18,000 hours validated recording hours~\cite{ardila2019common}. The MLS, a multilingual dataset, was derived from Librispeech~\cite{pratap2020mls}. Concerning linguistic diversity, the CMU Wilderness and MMS-Lab dataset, originating from the religious domain, covers nearly 1,000 languages~\cite{black2019cmu, pratap2023scaling}. Unlabeled dataset such as Libri-light has also been applied successfully to train self-supervised models such as HuBERT, wav2vec2, and WavLM~\cite{kahn2020libri,hsu2021hubert,baevski2020wav2vec,chen2022wavlm}. 

Despite the achievements with large-scale datasets, most public speech datasets available do not exceed 100,000 hours. In contrast, industry-utilized speech models are typically much more extensive. For instance, Whisper and Google's USM, have been trained with over 100,000 hours and up to 1,000,000 hours of data, respectively~\cite{radford2023robust, zhang2023google}. However, the details of the datasets used to train these models remain undisclosed, which makes it difficult to reproduce those models. Addressing the limitation of the lack of industry-scale large dataset, this paper presents YODAS (YouTube-Oriented Dataset of Audio and Speech)—a large-scale multilingual dataset, which comprises the following three subsets:


\begin{figure}[t]
  \centering
  \includegraphics[width=0.5\textwidth]{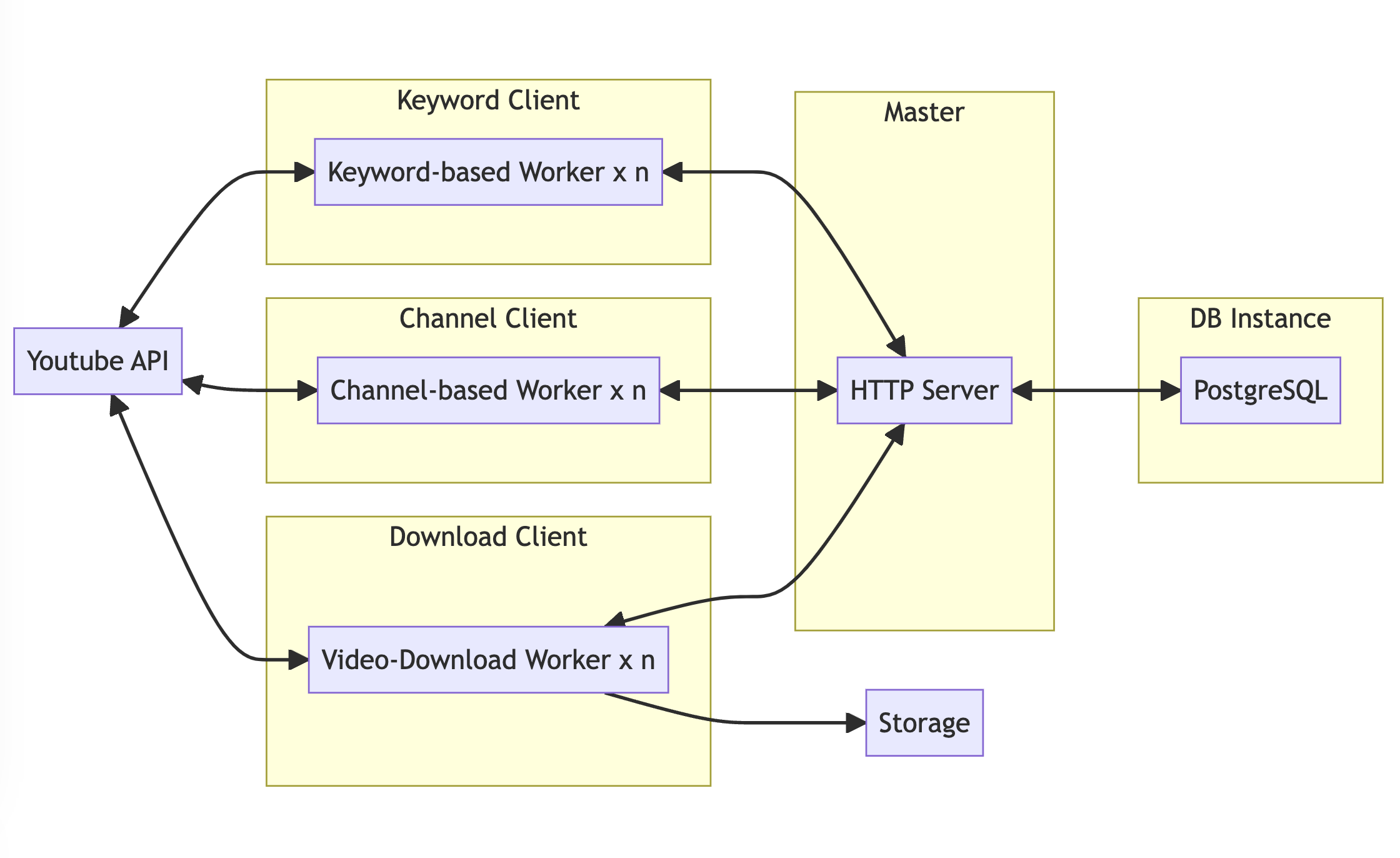}
  \caption{Diagram of our data collection architecture: It incorporates three types of clients: Keyword-based, Channel-based, and Download workers. Each worker fulfills specific tasks and interacts with both the master node and the YouTube platform. Additionally, the Download worker also handles the transfer of downloaded data to external storage.}
  \label{fig:arch}
\end{figure}


\begin{enumerate}[itemsep=2pt, parsep=2pt, topsep=2pt]
\item The \textit{manual} subset encompasses 86,400 hours of audio data paired with manual transcriptions.
\item The \textit{automatic} subset, containing 335,845 hours of audio data, is supplemented with automatic transcriptions.
\item The \textit{unlabeled} subset consists of 144,174 hours of raw audio data, devoid of any transcription.
\end{enumerate}

\begin{table*}[t]
  \centering
  \begin{tabular*}{\textwidth}{@{\extracolsep{\fill}}lrccccc}
    \toprule
    Dataset & \# Languages & Total Hours & Speech Type & Labeled & Public & License \\
    \midrule
    BABEL~\cite{gales2014speech} & 17 & 1k hours & Spontaneous & Yes & Yes & IARPA Babel License \\
    Common Voice~\cite{ardila2019common} & 112 & 18k hours & Read & Yes & Yes & CC-0 \\
    MLS~\cite{pratap2020mls} & 8 & 50.5k hours & Read & Yes & Yes & CC BY 4.0 \\
    FLEURS~\cite{conneau2023fleurs} & 102 & 1.4k hours & Read & Yes & Yes & CC BY 2.5 \\
    CMU Wilderness~\cite{black2019cmu} & 700 & 14k hours & Read & Yes & Yes & - \\
    MMS-Lab ~\cite{pratap2023scaling} & 1,107 & 44.7k hours & Read & Yes & No & - \\
    VoxLingua107~\cite{valk2021voxlingua107} & 107 &  6.6k hours & Spontaneous & Yes & Yes & CC BY 4.0 \\
    Librilight~\cite{kahn2020libri} & 1 & 60k hours & Read & No & Yes & CC BY 4.0 \\
    Whisper\cite{radford2023robust} & 97 & 680k hours & Unknown & Yes/No & No & - \\
    USM~\cite{zhang2023google} & 300 & 12M hours & Spontaneous & Yes/No & No & - \\
    \midrule
    YODAS (manual) & 140 & 86k hours &  Spontaneous & Yes & Yes & CC BY 3.0 \\
    YODAS (automatic) & 20 & 336k hours &  Spontaneous & Yes & Yes & CC BY 3.0 \\
    YODAS (unlabelled) & - & 144k hours &  Spontaneous & No & Yes & CC BY 3.0 \\
    \bottomrule
  \end{tabular*}
  \caption{A comparison of YODAS dataset with a few other large-scale multilingual datasets. Our YODAS dataset is the first public dataset to reach a scale of over 500k hours.}
\end{table*}
When used conjointly, the \textit{manual} and \textit{automatic} subsets offer a comprehensive resource of 480k hours for supervised model training. Meanwhile, all three subsets may be used in conjunction with the application of self-supervised learning techniques. 
It's worth noting that the combined subsets will result in an extensive corpus of over 560k hours by July 2023, and this amount will continue to grow over time.
This marks the first time a dataset of this scale with the Creative Commons license has been made publicly available. We plan to release it from the Huggingface datasets.

\section{Data Collection}

We designed our data collection architecture to fulfill two specific requirements. Firstly, the video content must be accompanied by a Creative Commons license. Secondly, the video should possess either an automatic subtitle or a manual subtitle as much as possible, although we also allow unlabeled videos. In order to meet these criteria, we devised a framework by improving upon an existing toolkit~\cite{takamichi2021jtubespeech}. Our framework is depicted in Figure~\ref{fig:arch}, which utilizes three distinct clients and a master node which we will discuss next.
Our data collection software will be open to the public for individual use or collaborative efforts.




\subsection{Keyword-based Client}

The principal challenge within our data gathering pipeline lies in the efficient identification of proper videos to download. This is achieved by implementing keyword-based crawling and combining it with YouTube's native filtering feature, allowing us to pinpoint a subset of relevant videos.

\begin{figure}[h]
  \centering
  \includegraphics[width=0.5\textwidth]{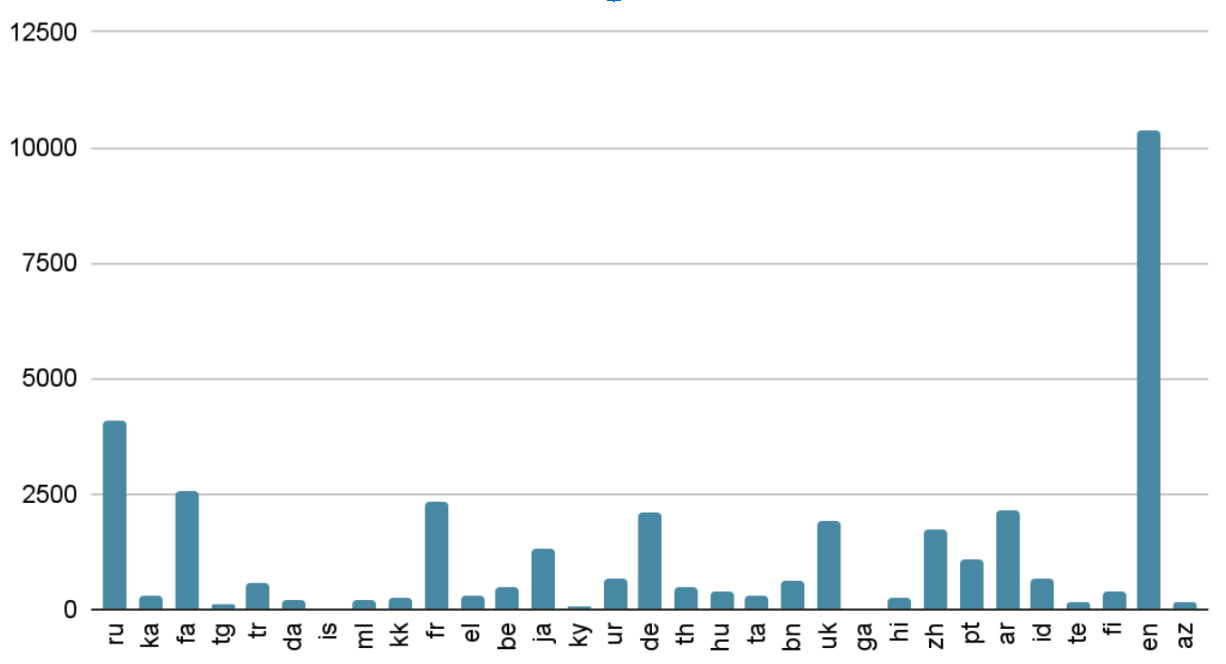}
  \caption{language distribution of unique query keywords used in one of our shards (i.e., worker).}
  \vspace*{-3mm}
  \label{fig:keyword}
\end{figure}

In the first phase, we construct a list of keywords by extracting unique keywords from the dumps of multilingual Wikipedia articles. Figure \ref{fig:keyword} presents the language distribution of unique query keywords within one of our data shards. As depicted by the figure, English commands the majority share. Rather than querying every keyword from this distribution, we prioritize those derived from less prominent languages trying to enhance the diversity of our video dataset. Subsequently, we initiate a keyword-based query on YouTube by appending appropriate flags, enforcing that the videos listed in the search results should (mostly) possess subtitles and adhere to the Creative Commons license. The naive HTTP GET request tends to yield a subset of videos that have the highest relevance to the supplied keywords. In order to capitalize on the number of available videos, we use AJAX to dynamically crawl lower-ranked videos, mimicking the process of scrolling down the search result page. 

\subsection{Channel-based Client}


The keyword-based crawling approach alone, unfortunately, is insufficient in scaling our dataset to the substantial size that we desire. This is due to the tendency of YouTube to repetitively present the same popular video subset in its search results rather than proposing new content. 

To mitigate this challenge and broaden our video exploration, we combine keyword-based crawling with a channel-based crawling strategy. For each video discovered during the keyword-based search, we extract the corresponding channel. This channel then serves as a means to identify all affiliated videos with ease. This method significantly aids in the discovery of videos that are otherwise less likely to appear using the keyword-centric approach. Most crucially, videos hosted within the same channel typically share similar licensing and subtitle characteristics, simplifying the process of identifying new videos that meet our specific criteria.

\subsection{Download Client}

The aforementioned workers solely undertake the task of identifying the correct video; however, the responsibility of downloading the video or its subtitles falls to the download worker. This worker perpetually monitors the database to ascertain whether a new video has been discovered by its predecessors. Upon identifying new videos, The download worker first retrieves the video, converting it into an audio format encoded at 1 channel and 24 kHz. Subsequently, it downloads the list of all available subtitles. Each video may either have multiple subtitles or none at all. The subtitles can be either manually uploaded by the user (manual subtitle) or automatically recognized by YouTube as enabled by the user (automatic subtitle). One significant challenge in this endeavor is determining the "correct" subtitle and the language of each video. Surprisingly, many videos possess multiple subtitles across diverse languages. We employ a heuristic method to identify the language and choose which subtitle to download. If the target video only has a unique manual subtitle with no other subtitles, it's highly probable this singular subtitle accurately reflects the language. We then proceed to download this subtitle and assign the language ID. Similarly, if the target video only has one automatic subtitle, we consider it to be accurate and proceed to download this subtitle. However, when a video has more than one manual or automatic subtitle with conflicting language IDs, the task of identifying the correct language becomes complicated. In such cases, we forego downloading the subtitles and leaving the video unlabeled. Although we attempt to identify the language using Language Identification (LID) tools, the results are not significantly successful. Therefore, we earmark this identification task for future work.

\subsection{Master Node}


In addition to those clients, we deploy a master node to monitor the overall progress. This node is connected to a PostgreSQL database and hosts an HTTP server that accepts GET/POST HTTP requests from each worker. The master node manages all resources—be it keywords, channels, or videos—each marked with one of three states: not-started, being processed, or done. The 'being processed' state serves to prevent simultaneous downloads of the same resource by separate workers. All the workers typically function as HTTP clients, querying the master node to request the next available resource. Once the resource has been resolved (i.e., downloaded), these workers mark it as done. 



\section{Analysis}

\begin{figure*}[t]
  \centering
  \includegraphics[width=\textwidth]{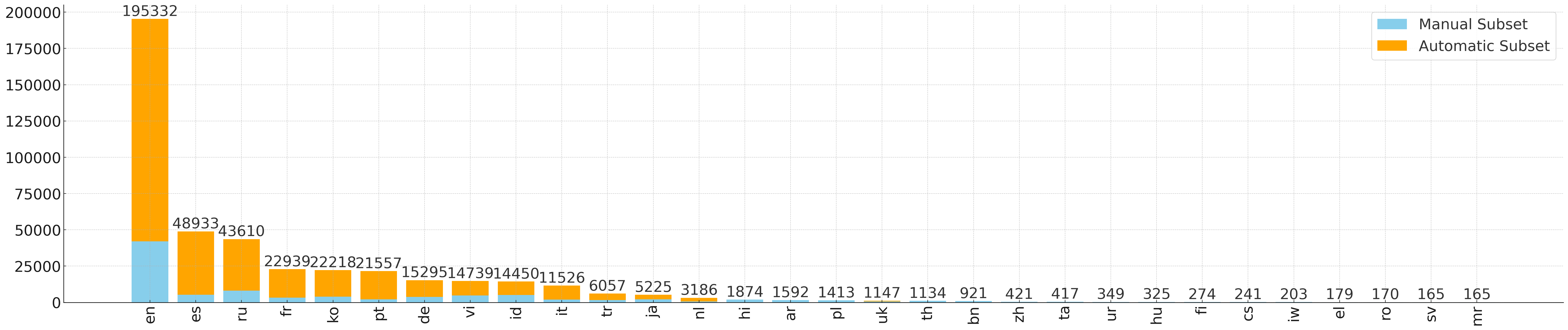}
  \caption{Total duration (measured in hours) in the manual and automatic subset. The lower-blue bar shows the duration of the manual subset, the top-orange bar indicates the automatic subset. The combined duration is illustrated on top of each bar.}
  \label{fig:manual_auto_hour}
\end{figure*}

\subsection{Overview}

As described in the previous section, the YODAS dataset was segmented into three subsets, namely: \textit{Manual}, \textit{Automatic}, and \textit{Unlabeled}. Videos within the manual subset are characterized by user-uploaded, (possibly) manually generated subtitles, while those included in the automatic subset have associated automated subtitles. Conversely, the unlabeled subset consists of videos devoid of subtitles, primarily due to our current inability to accurately identify the language. 

\subsection{Speech Analysis}

Table \ref{tab:video_duration} presents the key statistics concerning the distribution of raw video durations and utterance durations (shown in parentheses) within our datasets. Notably, the raw video duration of the automatic subset exhibits a notably higher mean duration and standard deviation compared to the other two datasets. Conversely, the average duration of utterances and its standard deviation are notably lower in the automatic subset as compared to the manual subset. This is because YouTube tends to chunk speech into small segments as we will discuss further in the next subsection. In total, we have compiled an extensive dataset comprising 86,000 hours for the manual subset, 336,000 hours for the automatic subset, and an additional 144,000 hours for the unlabeled subset.


\begin{table}[h!]
  \centering
  \begin{tabular}{lccc}
    \toprule
    & Manual & Automatic & Unlabeled \\
    \midrule
    Mean & 0.15h (5.6s) & 0.23h (3.2s) & 0.15h (-)\\
    Std & 0.35h (8.9s)& 0.37h (1.6s)& 0.25h (-)\\
    Min & 0.00h (0.0s) & 0.00h (0.1s) & 0.00h (-)\\
    Max & 24.9h (42.1s) & 24.9h (87.7s) & 24.9h (-)\\
    \bottomrule
  \end{tabular}
 \caption{Descriptive statistics of raw video duration distribution (measured in hours) and utterance duration distribution in parentheses (measured in seconds) in three subsets.}
   \label{tab:video_duration}
\end{table}



Next, the language distribution of the manual subset and the automatic subset is illustrated in Figure~\ref{fig:manual_auto_hour}. This figure presents the distribution of the top 30 languages out of a total of 140 evaluated. As anticipated, English (en) emerges as the most prevalently used language, with Spanish (es) and Russian (ru) occupying the second and third positions, respectively, when assessed based on duration. Although the language distribution trend appears similar between the automatic and manual subsets, the automatic subset has only a very limited number of languages (14 languages) compared with the manual subset (140 languages).







\subsection{Text Analysis}

The top 10 writing systems in our dataset are Latin, Cyrillic, CJK, Hiragana, Greek, Devanagari, Hangul,  Malayalam, Katakana, and Arabic. It largely corresponds to the aforementioned language distribution, with the Latin alphabet appearing as the most frequently used writing system. The Cyrillic script, originating from Russian language videos, also features prominently within our dataset.



\begin{table}[h!]
  \centering
  \begin{tabular}{lccc}
    \toprule
    & Manual & Automatic\\
     \midrule
    Mean & 58.2 & 33.5\\
    Std & 27.6 & 8.6\\
    Min & 0.0 & 0.0\\
    Max & 588 & 44\\
    \bottomrule
  \end{tabular}
  \caption{Descriptive statistics of subtitle transcription measured in the number of characters in two subsets.}
    \label{tab:text_length}
\end{table}

Table~\ref{tab:text_length} is the comparison of character length distribution from the manual subset and automatic subset. The manual subset tends to have a larger number of characters per utterance and have a larger variance. Conversely, the automatic subset has an even distribution where most utterances are short and have little variance. This is because the automatic subtitle frequently divides long utterances into small chunks as Table~\ref{tab:sample} indicates, this splitting might be a feature to help viewers to follow subtitles easier.

\begin{table}[h!]
  \centering
  \begin{tabular}{ll}
    \toprule
    utt id & automatic transcription \\
    \midrule
    00682 & if you're trying to do something in your \\
    00683 & community and you're spending your money \\
    00684 & public money or somebody's money to \\
    00685 & really do something that makes a \\
    \bottomrule
  \end{tabular}
\caption{A sample of the automatic transcriptions, one individual utterance is usually divided into multiple small chunks.}
  \label{tab:sample}
\end{table}




\section{Experiment}


The YODAS dataset offers a versatile resource that can be employed for a variety of tasks including supervised training, weakly-labeled supervised training, and self-supervised training. In this work, we focus specifically on the use of the dataset for the monolingual speech recognition task.

\subsection{Speech-Text Alignment}
\label{sec:alignment}

The raw dataset, as collected, presents considerable noise with regard to speech-text alignment, suggesting that its unfiltered usage might be inappropriate. There are instances where neither the manual nor automatic subtitles accurately represent the underlying spoken discourse. For instance, subtitles occasionally serve as concise descriptors of the current scene in a video, annotating elements such as laughter or musical segments, rather than transcribing the actual spoken dialogue. Our heuristics to decide the language based on the list of subtitles might also introduce errors, those errors possibly arising from user inaccuracies or misidentifications in YouTube's language detection system. 

\begin{figure}[h]
  \centering
  \includegraphics[width=0.5\textwidth]{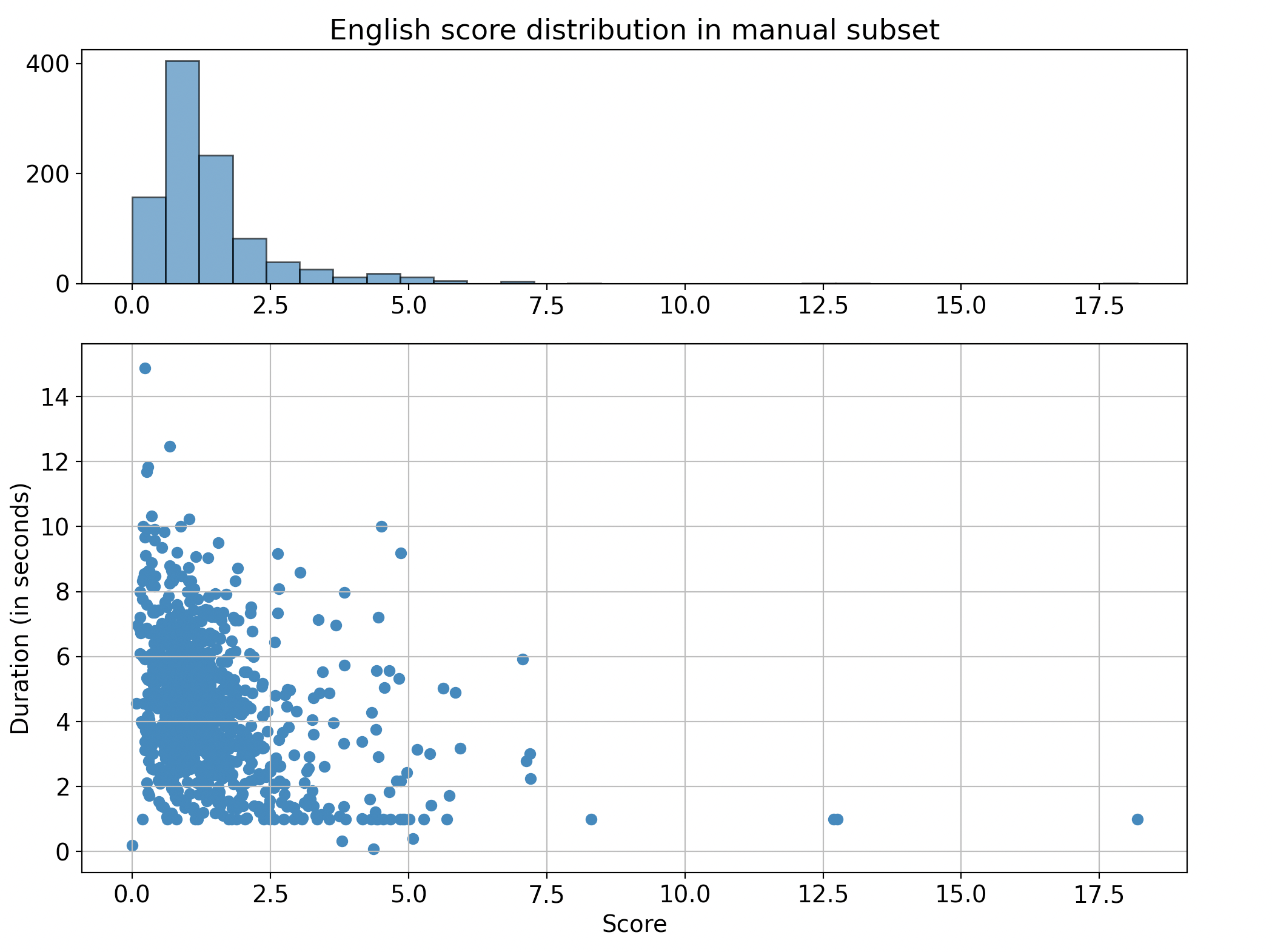}
  \caption{the score histogram and scatter plot of the relationship between the duration and the alignment in the manual subset.}
  \label{fig:manual_score_dist}
\end{figure}

\begin{figure}[h]
  \centering
  \includegraphics[width=0.5\textwidth]{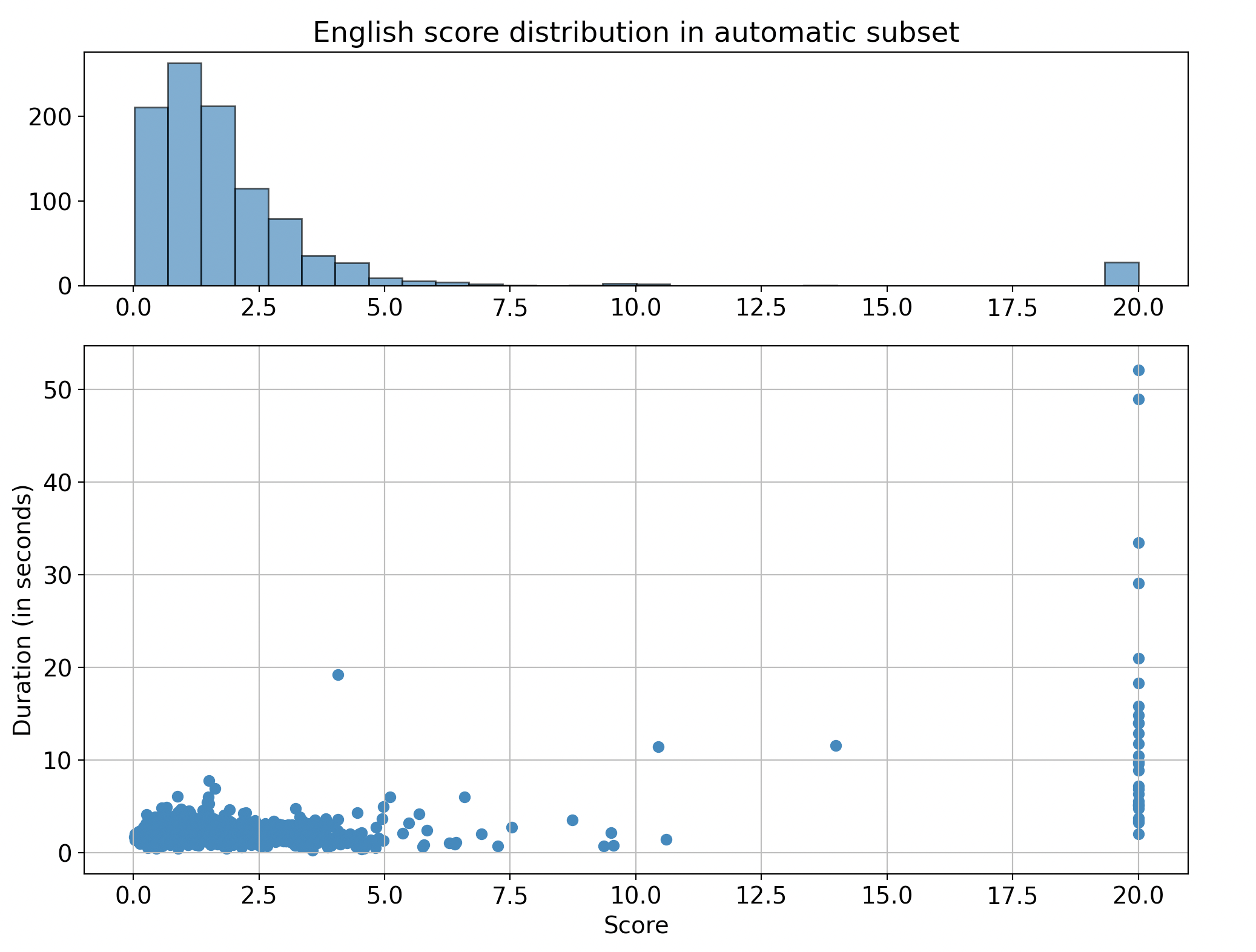}
  \caption{the score histogram and scatter plot between the duration and the alignment score in the automatic subset.}
  \label{fig:auto_score_dist}
\end{figure}

To filter the dataset, we first apply the speech-text alignment. In particular, we use a pre-trained acoustic model to score every utterance in the audio and only consider using the high-scoring utterance pairs~\cite{li2020universal}. The score is obtained from the CTC loss where a lower value (loss) implies a better alignment~\cite{graves2014towards,kurzinger2020ctc}. Figure~\ref{fig:manual_score_dist} presents a scatter plot depicting the relationship between the duration and alignment score of 1,000 utterances randomly sampled from the \textit{manual} subset. From the plot, it is evident that while there are occasional outliers with poor alignment scores such as 18.0, the majority of utterances exhibit a duration of less than 10 seconds and possess an alignment score superior to 5.  Conversely, Figure \ref{fig:auto_score_dist} portrays a scatter plot derived from the \textit{automatic} subset. This plot reveals a significant proportion of misaligned utterances. It should be noted that for analytical purposes, all scores exceeding 20 have been capped at 20. These misaligned utterances typically display a considerable duration, often extending to as long as 50 seconds, and are predominantly attributed to music or background noise. 

In the subsequent experiment, we employ an alignment threshold to exclude utterances with scores worse than \textbf{2.0}. From the refined subset, a sample of 1,000,000 utterances (at most) is randomly selected to constitute the training set for each language, while a separate, smaller selection of 1,000 utterances is assigned to the testing dataset.


\subsection{Baseline}

\begin{table*}[t]
  \centering
  \begin{tabular*}{\textwidth}{@{\extracolsep{\fill}}lrrrrrrrrrrrrrrrrrr}
    \toprule
     language & ell & nld & hun & pol & por & cmn & ind & jpn & tur & ita & deu & fra & spa & rus & eng & average\\
    \midrule
    CER ($\downarrow$) & 8.3 & 8.4 & 6.2 & 6.4 & 7.1 & 12.5 & 10.2 & 14.7 & 8.7 & 8.6 & 10.1 & 12.9 & 9.8 & 12.8 & 12.9 & 9.97 \\
    \bottomrule
  \end{tabular*}
  \caption{monolingual speech recognition performance on the top-15 languages (ordered by the duration) from the manual subset. The evaluation is done by using character error rate (CER) where a lower number indicates a better performance.}
    \label{tab:main_result}
\end{table*}

We build simple monolingual baseline models for the top-25 languages in the manual subset. Our model is based on the pre-trained XLSR representations~\cite{babu2021xls}, where we have a linear layer randomly initialized on top of the pre-trained representations, which is then optimized with the CTC loss~\cite{graves2014towards}. The preparation is done by using ESPnet~\cite{watanabe2018espnet} and s3prl~\cite{yang21c_interspeech}. The subword vocabulary is prepared with BPE using SentencePiece~\cite{sennrich-etal-2016-neural,kudo2018subword}, where we use 300 as the vocabulary size for most languages except for CJK languages where we use 5000 for Mandarin and 3000 for Japanese. For simplicity, we do not perform speech augmentation such as SpecAugment~\cite{park19e_interspeech} and Speed Perturbation~\cite{ko2015audio}. The acoustic model is optimized with the AdamW optimizer with a fixed learning rate of 0.0001~\cite{loshchilov2017decoupled}. The decoding is done greedily without any language models.


\subsection{Results}


Table.\ref{tab:main_result} displays the testing outcomes for the top-15 languages, measured by the Character Error Rate (CER). The respective CER for each language spans from 6 to 15. The best performance is recorded for Hungarian, with a CER of 6.2, while Japanese exhibits the least performance with a CER of 14.7. The average CER across all languages is 9.97. We observe that languages possessing a larger BPE vocabulary size, such as Mandarin (cmn) and Japanese (jpn), tend to correspond with higher character error rates (Mandarin has a CER of 12.5 and Japanese has a CER of 14.7). Conversely, languages that adhere to more straightforward spelling rules generally exhibit lower character error rates. For example, the writing system in Hungarian is mostly phonemic and achieves the lowest CER 6.2 in our experiment.

\begin{table}[h]
  \centering
  \begin{tabular}{lrrrr}
    \toprule
     & CER ($\downarrow$) & Add ($\downarrow$) & Del ($\downarrow$) & Sub ($\downarrow$) \\
    \midrule
    Manual & 14.9 & 3.2 & 7.5 & 4.2 \\
    Automatic & 32.3 & 1.6 & 26.6 & 4.1\\
    \bottomrule
  \end{tabular}
  \caption{A comparison of the speech models trained with the manual subset and the automatic subset. We demonstrate both the CER and its error decomposition of Addition (Add), Deletion (Del), and Substitution (Sub).}
    \label{tab:manual_vs_auto}
\end{table}


\begin{figure}[h]
  \centering
  \includegraphics[width=0.5\textwidth]{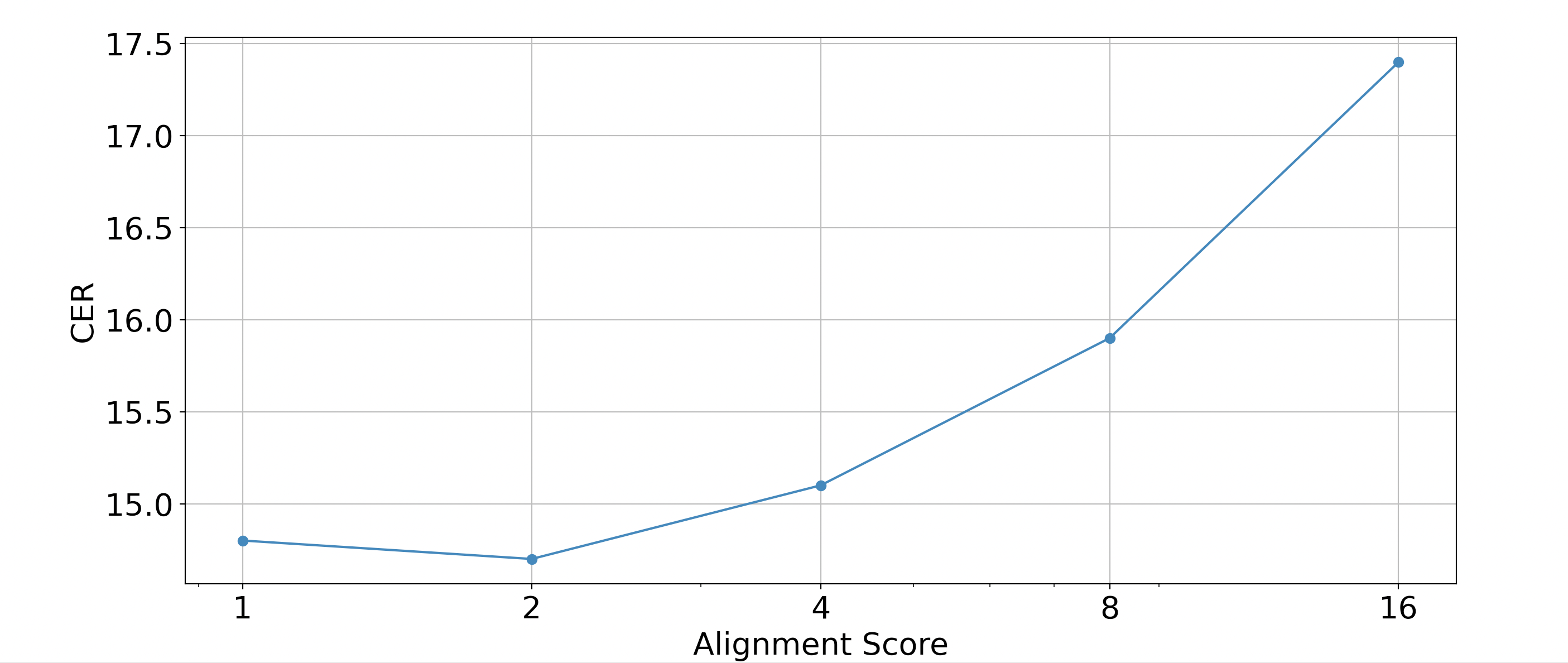}
  \caption{the relationship between the alignment score and its performance on the speech recognition task.}
  \label{fig:alignment_trend}
\end{figure}

We subsequently analyzed the quality of the dataset across both the manual and automatic subsets by training independent models solely on each subset. For a balanced comparison, 100,000 utterances were randomly selected from each subset, post application of the 2.0 score filter, as introduced in Section \ref{sec:alignment}. This comparison was solely conducted within the English subset. The findings, as displayed in Table \ref{tab:manual_vs_auto}, reveal that models trained on the manual subset yield significantly superior performance compared to those trained on the automatic subset. Further analysis indicates the primary cause of this discrepancy was the deletion error. The automatic subset presented a notably high deletion error rate of 26.6, whereas the manual subset recorded a markedly lower rate of 7.0. These findings align with previous research, which has indicated that the use of automatically-generated transcripts tends to undermine system performance \cite{radford2023robust, ghorbani2021scaling}. Consequently, these results underscore the importance of prioritizing the utilization of the manual subset over the automatic subset in the training of models.

\begin{figure}[h]
  \centering
  \includegraphics[width=0.5\textwidth]{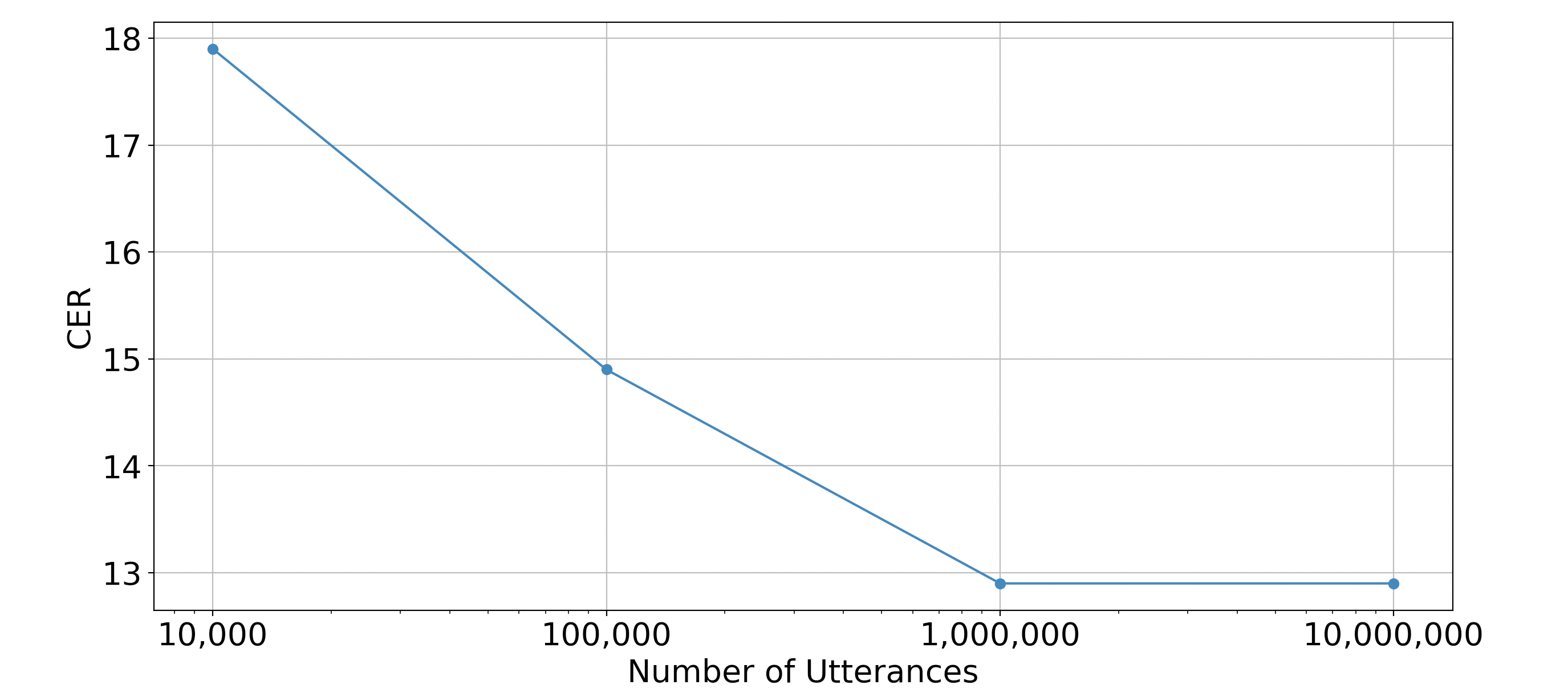}
  \caption{the relationship between the size of the training dataset and its performance on the speech recognition task.}
  \label{fig:size_trend}
\end{figure}


In the previous experiments, an alignment score threshold of 2.0 was implemented as a cut-off. To investigate how altering this threshold might impact performance, we performed an experiment with varied threshold values, ranging from 1 to 16. For analytical purposes, scores exceeding 16 were normalized to 16. We conducted this analysis using 100,000 utterances (160 hours) from the manual subset with the same testing set. The outcome of this exercise is illustrated in Figure \ref{fig:alignment_trend}. It is evident that the setting of the alignment threshold is critical to model performance where raising the threshold from 16 to 2 results in consistent performance improvement. However, further tightening of the threshold from 2.0 to 1.0 caused a minor degradation in performance, from a CER of 14.7 to 14.8. This slight decrease may be attributed to the fact that utterances in the training set with an alignment score of 1.0 tend to be shorter and comprise fewer words than those within the subset with a score of 2.0. These findings imply the significance of judiciously selecting the alignment threshold when training models. Although a lower threshold might seem intuitively beneficial, it may inadvertently exclude longer, richer utterances, thereby potentially impacting performance.


Finally, we investigated the impact of modifying the size of the training dataset, ranging from 10,000 to 10 million utterances (16 to 16k hours). All utterances were randomly selected from the English manual subset, adhering to an alignment threshold of 2.0. The result is depicted in Table \ref{fig:size_trend}. The results demonstrate that increasing the quantity of utterances consistently enhances model performance. Interestingly, the model trained with 1 million utterances and the model trained with 10 million utterances exhibit a similar CER. This phenomenon could potentially be attributed to the simplistic architecture we employed - namely a linear model built upon pre-trained features. Such an architecture may not be fully equipped to leverage the expanded dataset. 


\section{Conclusion}

In this study, we presented the YODAS dataset, a comprehensive, multilingual dataset compiled from the YouTube platform. We delineated our data collection pipeline and provided preliminary analyses and baseline models based on the dataset. We anticipate that the YODAS dataset will serve as a valuable resource for the speech research community

\section{Acknowledgements}
We use PSC Bridges2 and NCSA Delta via ACCESS allocation CIS210014, supported by National Science Foundation grants \#2138259, \#2138286, \#2138307, \#2137603, and \#2138296.

\bibliographystyle{IEEEbib}
\bibliography{strings,refs}

\begin{thebibliography}{10}

\bibitem{graves2006connectionist}
Alex Graves, Santiago Fern{\'a}ndez, Faustino Gomez, and J{\"u}rgen
  Schmidhuber,
\newblock ``Connectionist temporal classification: labelling unsegmented
  sequence data with recurrent neural networks,''
\newblock in {\em Proceedings of the 23rd international conference on Machine
  learning}, 2006, pp. 369--376.

\bibitem{collobert2016wav2letter}
Ronan Collobert, Christian Puhrsch, and Gabriel Synnaeve,
\newblock ``Wav2letter: an end-to-end convnet-based speech recognition
  system,''
\newblock {\em arXiv preprint arXiv:1609.03193}, 2016.

\bibitem{graves2013speech}
Alex Graves, Abdel-rahman Mohamed, and Geoffrey Hinton,
\newblock ``Speech recognition with deep recurrent neural networks,''
\newblock in {\em 2013 IEEE international conference on acoustics, speech and
  signal processing}. Ieee, 2013, pp. 6645--6649.

\bibitem{sutskever2014sequence}
Ilya Sutskever, Oriol Vinyals, and Quoc~V Le,
\newblock ``Sequence to sequence learning with neural networks,''
\newblock {\em Advances in neural information processing systems}, vol. 27,
  2014.

\bibitem{prabhavalkar2023end}
Rohit Prabhavalkar, Takaaki Hori, Tara~N Sainath, Ralf Schl{\"u}ter, and Shinji
  Watanabe,
\newblock ``End-to-end speech recognition: A survey,''
\newblock {\em arXiv preprint arXiv:2303.03329}, 2023.

\bibitem{baevski2020wav2vec}
Alexei Baevski, Yuhao Zhou, Abdelrahman Mohamed, and Michael Auli,
\newblock ``wav2vec 2.0: A framework for self-supervised learning of speech
  representations,''
\newblock {\em Advances in neural information processing systems}, vol. 33, pp.
  12449--12460, 2020.

\bibitem{hsu2021hubert}
Wei-Ning Hsu, Benjamin Bolte, Yao-Hung~Hubert Tsai, Kushal Lakhotia, Ruslan
  Salakhutdinov, and Abdelrahman Mohamed,
\newblock ``Hubert: Self-supervised speech representation learning by masked
  prediction of hidden units,''
\newblock {\em IEEE/ACM Transactions on Audio, Speech, and Language
  Processing}, vol. 29, pp. 3451--3460, 2021.

\bibitem{mohamed2022self}
Abdelrahman Mohamed, Hung-yi Lee, Lasse Borgholt, Jakob~D Havtorn, Joakim Edin,
  Christian Igel, Katrin Kirchhoff, Shang-Wen Li, Karen Livescu, Lars
  Maal{\o}e, et~al.,
\newblock ``Self-supervised speech representation learning: A review,''
\newblock {\em IEEE Journal of Selected Topics in Signal Processing}, 2022.

\bibitem{gales2014speech}
Mark~JF Gales, Kate~M Knill, Anton Ragni, and Shakti~P Rath,
\newblock ``Speech recognition and keyword spotting for low-resource languages:
  Babel project research at cued,''
\newblock in {\em Fourth International workshop on spoken language technologies
  for under-resourced languages (SLTU-2014)}. International Speech
  Communication Association (ISCA), 2014, pp. 16--23.

\bibitem{ardila2019common}
Rosana Ardila, Megan Branson, Kelly Davis, Michael Henretty, Michael Kohler,
  Josh Meyer, Reuben Morais, Lindsay Saunders, Francis~M Tyers, and Gregor
  Weber,
\newblock ``Common voice: A massively-multilingual speech corpus,''
\newblock {\em arXiv preprint arXiv:1912.06670}, 2019.

\bibitem{pratap2020mls}
Vineel Pratap, Qiantong Xu, Anuroop Sriram, Gabriel Synnaeve, and Ronan
  Collobert,
\newblock ``Mls: A large-scale multilingual dataset for speech research,''
\newblock {\em arXiv preprint arXiv:2012.03411}, 2020.

\bibitem{black2019cmu}
Alan~W Black,
\newblock ``Cmu wilderness multilingual speech dataset,''
\newblock in {\em ICASSP 2019-2019 IEEE International Conference on Acoustics,
  Speech and Signal Processing (ICASSP)}. IEEE, 2019, pp. 5971--5975.

\bibitem{pratap2023scaling}
Vineel Pratap, Andros Tjandra, Bowen Shi, Paden Tomasello, Arun Babu, Sayani
  Kundu, Ali Elkahky, Zhaoheng Ni, Apoorv Vyas, Maryam Fazel-Zarandi, et~al.,
\newblock ``Scaling speech technology to 1,000+ languages,''
\newblock {\em arXiv preprint arXiv:2305.13516}, 2023.

\bibitem{kahn2020libri}
Jacob Kahn, Morgane Rivi{\`e}re, Weiyi Zheng, Evgeny Kharitonov, Qiantong Xu,
  Pierre-Emmanuel Mazar{\'e}, Julien Karadayi, Vitaliy Liptchinsky, Ronan
  Collobert, Christian Fuegen, et~al.,
\newblock ``Libri-light: A benchmark for asr with limited or no supervision,''
\newblock in {\em ICASSP 2020-2020 IEEE International Conference on Acoustics,
  Speech and Signal Processing (ICASSP)}. IEEE, 2020, pp. 7669--7673.

\bibitem{chen2022wavlm}
Sanyuan Chen, Chengyi Wang, Zhengyang Chen, Yu~Wu, Shujie Liu, Zhuo Chen, Jinyu
  Li, Naoyuki Kanda, Takuya Yoshioka, Xiong Xiao, et~al.,
\newblock ``Wavlm: Large-scale self-supervised pre-training for full stack
  speech processing,''
\newblock {\em IEEE Journal of Selected Topics in Signal Processing}, vol. 16,
  no. 6, pp. 1505--1518, 2022.

\bibitem{radford2023robust}
Alec Radford, Jong~Wook Kim, Tao Xu, Greg Brockman, Christine McLeavey, and
  Ilya Sutskever,
\newblock ``Robust speech recognition via large-scale weak supervision,''
\newblock in {\em International Conference on Machine Learning}. PMLR, 2023,
  pp. 28492--28518.

\bibitem{zhang2023google}
Yu~Zhang, Wei Han, James Qin, Yongqiang Wang, Ankur Bapna, Zhehuai Chen, Nanxin
  Chen, Bo~Li, Vera Axelrod, Gary Wang, et~al.,
\newblock ``Google usm: Scaling automatic speech recognition beyond 100
  languages,''
\newblock {\em arXiv preprint arXiv:2303.01037}, 2023.

\bibitem{conneau2023fleurs}
Alexis Conneau, Min Ma, Simran Khanuja, Yu~Zhang, Vera Axelrod, Siddharth
  Dalmia, Jason Riesa, Clara Rivera, and Ankur Bapna,
\newblock ``Fleurs: Few-shot learning evaluation of universal representations
  of speech,''
\newblock in {\em 2022 IEEE Spoken Language Technology Workshop (SLT)}. IEEE,
  2023, pp. 798--805.

\bibitem{valk2021voxlingua107}
J{\"o}rgen Valk and Tanel Alum{\"a}e,
\newblock ``Voxlingua107: a dataset for spoken language recognition,''
\newblock in {\em 2021 IEEE Spoken Language Technology Workshop (SLT)}. IEEE,
  2021, pp. 652--658.

\bibitem{takamichi2021jtubespeech}
Shinnosuke Takamichi, Ludwig K{\"u}rzinger, Takaaki Saeki, Sayaka Shiota, and
  Shinji Watanabe,
\newblock ``Jtubespeech: corpus of japanese speech collected from youtube for
  speech recognition and speaker verification,''
\newblock {\em arXiv preprint arXiv:2112.09323}, 2021.

\bibitem{li2020universal}
Xinjian Li, Siddharth Dalmia, Juncheng Li, Matthew Lee, Patrick Littell, Jiali
  Yao, Antonios Anastasopoulos, David~R Mortensen, Graham Neubig, Alan~W Black,
  et~al.,
\newblock ``Universal phone recognition with a multilingual allophone system,''
\newblock in {\em ICASSP 2020-2020 IEEE International Conference on Acoustics,
  Speech and Signal Processing (ICASSP)}. IEEE, 2020, pp. 8249--8253.

\bibitem{graves2014towards}
Alex Graves and Navdeep Jaitly,
\newblock ``Towards end-to-end speech recognition with recurrent neural
  networks,''
\newblock in {\em International conference on machine learning}. PMLR, 2014,
  pp. 1764--1772.

\bibitem{kurzinger2020ctc}
Ludwig K{\"u}rzinger, Dominik Winkelbauer, Lujun Li, Tobias Watzel, and Gerhard
  Rigoll,
\newblock ``Ctc-segmentation of large corpora for german end-to-end speech
  recognition,''
\newblock in {\em International Conference on Speech and Computer}. Springer,
  2020, pp. 267--278.

\bibitem{babu2021xls}
Arun Babu, Changhan Wang, Andros Tjandra, Kushal Lakhotia, Qiantong Xu, Naman
  Goyal, Kritika Singh, Patrick von Platen, Yatharth Saraf, Juan Pino, et~al.,
\newblock ``Xls-r: Self-supervised cross-lingual speech representation learning
  at scale,''
\newblock {\em arXiv preprint arXiv:2111.09296}, 2021.

\bibitem{watanabe2018espnet}
Shinji Watanabe, Takaaki Hori, Shigeki Karita, Tomoki Hayashi, Jiro Nishitoba,
  Yuya Unno, Nelson {Enrique Yalta Soplin}, Jahn Heymann, Matthew Wiesner,
  Nanxin Chen, Adithya Renduchintala, and Tsubasa Ochiai,
\newblock ``{ESPnet}: End-to-end speech processing toolkit,''
\newblock in {\em Proceedings of Interspeech}, 2018, pp. 2207--2211.

\bibitem{yang21c_interspeech}
Shu wen Yang, Po-Han Chi, Yung-Sung Chuang, Cheng-I~Jeff Lai, Kushal Lakhotia,
  Yist~Y. Lin, Andy~T. Liu, Jiatong Shi, Xuankai Chang, Guan-Ting Lin,
  Tzu-Hsien Huang, Wei-Cheng Tseng, Ko~tik Lee, Da-Rong Liu, Zili Huang, Shuyan
  Dong, Shang-Wen Li, Shinji Watanabe, Abdelrahman Mohamed, and Hung yi~Lee,
\newblock ``{SUPERB: Speech Processing Universal PERformance Benchmark},''
\newblock in {\em Proc. Interspeech 2021}, 2021, pp. 1194--1198.

\bibitem{sennrich-etal-2016-neural}
Rico Sennrich, Barry Haddow, and Alexandra Birch,
\newblock ``Neural machine translation of rare words with subword units,''
\newblock in {\em Proceedings of the 54th Annual Meeting of the Association for
  Computational Linguistics (Volume 1: Long Papers)}, Berlin, Germany, Aug.
  2016, pp. 1715--1725, Association for Computational Linguistics.

\bibitem{kudo2018subword}
Taku Kudo,
\newblock ``Subword regularization: Improving neural network translation models
  with multiple subword candidates,''
\newblock {\em arXiv preprint arXiv:1804.10959}, 2018.

\bibitem{park19e_interspeech}
Daniel~S. Park, William Chan, Yu~Zhang, Chung-Cheng Chiu, Barret Zoph, Ekin~D.
  Cubuk, and Quoc~V. Le,
\newblock ``{SpecAugment: A Simple Data Augmentation Method for Automatic
  Speech Recognition},''
\newblock in {\em Proc. Interspeech 2019}, 2019, pp. 2613--2617.

\bibitem{ko2015audio}
Tom Ko, Vijayaditya Peddinti, Daniel Povey, and Sanjeev Khudanpur,
\newblock ``Audio augmentation for speech recognition,''
\newblock in {\em Sixteenth annual conference of the international speech
  communication association}, 2015.

\bibitem{loshchilov2017decoupled}
Ilya Loshchilov and Frank Hutter,
\newblock ``Decoupled weight decay regularization,''
\newblock {\em arXiv preprint arXiv:1711.05101}, 2017.

\bibitem{ghorbani2021scaling}
Behrooz Ghorbani, Orhan Firat, Markus Freitag, Ankur Bapna, Maxim Krikun,
  Xavier Garcia, Ciprian Chelba, and Colin Cherry,
\newblock ``Scaling laws for neural machine translation,''
\newblock {\em arXiv preprint arXiv:2109.07740}, 2021.

\end{thebibliography}

\end{document}